\documentclass[runningheads]{llncs}
\usepackage[T1]{fontenc}
\usepackage{graphicx}
\usepackage{makecell}
\usepackage{multirow}
\usepackage{multicol}
\usepackage{booktabs}
\usepackage{amsmath,amssymb,amsfonts}
\usepackage[table]{xcolor}  
\usepackage{pifont}
\usepackage{xcolor}  % If you want to color it
\definecolor{myRed}{RGB}{195,10,10}
\definecolor{myGreen}{RGB}{55,149,73}
\definecolor{myBlue}{RGB}{5,5,60}

\definecolor{Gray}{gray}{0.9}
\usepackage{marvosym} % For \Letter symbol

\begin{document}
\title{
Reasoning Text-to-Video Retrieval for Operating Room Clips via Action-Driven Digital Twins
}
\titlerunning{Reasoning Text-to-Video Retrieval}
\author{Yiqing Shen \and Hao Ding \and Mathias Unberath\textsuperscript{(\Letter)}}

\authorrunning{Y. Shen et al.}
\institute{Johns Hopkins University, Baltimore, MD, USA\\
\email{\{yshen92,unberath\}@jhu.edu}}
\maketitle              % typeset the header of the contribution

\begin{abstract}

Text-to-video retrieval in operating rooms (OR) is an enabling technology for OR safety, as it allows stakeholders to retrieve and inspect recordings of specific events. However, because the most safety critical events may not follow the common structure, to unlock its full potential text-to-video retrieval must be able to handle implicit queries that require reasoning to identify the right videos (\textit{e}.\textit{g}., ``\textit{the step right before clipping}''). However, existing methods rely on global embeddings that cannot reason over such queries.
We propose OR\textsuperscript{3}, a text-to-video retrieval method that converts clips into action-driven digital twins (ActDTs), grouping concurrent subject-action-object triplets under non-overlapping temporal intervals.
Moreover, rather than cross-modal matching through paired encoders, OR\textsuperscript{3} performs imagination-based retrieval where an LLM generates hypothetical ActDTs from queries.
This enables intra-modal matching via a single encoder trained with ActDT-tailored hard negatives.
Finally, evidence-grounded refinement revises imagined ActDTs based on discrepancies with top candidates to capture procedure-specific patterns.
We construct a benchmark from MM-OR with 276 implicit queries across four reasoning categories over 386 clips from robotic knee procedures.
OR\textsuperscript{3} achieves 57.6\% R@1 and 77.3\% R@5, outperforming the strongest baseline.
These results demonstrate that OR\textsuperscript{3} enables fine-grained discrimination between visually similar OR video clips through temporal action reasoning.


\end{abstract}

\section{Introduction}

Operating rooms (ORs) can produce thousands of hours of video from room-mounted cameras, yet users typically need only short clips that last seconds to minutes~\cite{levin2021surgical,mmor}.
For example, surgical trainees may search for a specific dissection technique~\cite{youssef2023learning,henning2025step}, while quality teams may need to locate moments of miscommunication~\cite{gruter2023video,shen2025operating}.
Because entire OR videos typically span multiple hours, clip-level text-to-video retrieval is a natural formulation where the OR videos are first segmented by surgical phase and steps~\cite{yu2023live}.
Unlike conventional text-to-video retrieval where queries explicitly describe visual content, OR retrieval must handle implicit queries demanding reasoning (\textit{e}.\textit{g}., ``\textit{the step right before clipping}'' or ``\textit{the action that caused bleeding}'')~\cite{shen2025operating,shen2025temporally}.
Moreover, the OR video retrieval demands fine-grained discrimination because these clips from the same procedure appear visually near-identical, differing only in actions performed and entity interactions~\cite{sharma2025fine}.

Existing text-to-video retrieval methods learn aligned text and video embeddings based on global similarity, but offer no mechanism to reason over implicit queries~\cite{zhang2025text,wang2021t2vlad}.
Recent work on reasoning text-to-video retrieval addresses this by converting videos into digital twins (DTs) over which large language models (LLMs) can reason during retrieval~\cite{reasont2v,shen2025operating}.
However, its object-centric DTs focus on per-frame entities and their visual attributes, which fail to discriminate between consecutive OR clips that share identical objects, but differ in actions and state transitions~\cite{reasont2v}.
Moreover, its cross-modal matching between DT representation and query through paired encoders struggles to bridge the semantic gap between abstract implicit queries and concrete DT representation in OR contexts~\cite{che2024cross}.

We propose OR\textsuperscript{3} (operating room reasoning retrieval), a reasoning retrieval method for OR clips.
OR\textsuperscript{3} represents each clip as an action-driven digital twin (ActDT), a two-level structure that groups concurrent object interactions under non-overlapping temporal intervals.
This action-centric representation captures state transitions and their dynamics, enabling discrimination between visually near-identical clips.
Rather than matching queries against ActDTs through cross-modal paired encoders, OR\textsuperscript{3} performs imagination-based retrieval by generating a hypothetical ActDT from the query itself.
This casts retrieval as intra-modal matching and eliminates the semantic gap between abstract implicit queries and concrete ActDTs.
When discrepancies emerge between imagined and actual ActDTs, evidence-grounded refinement revises the imagined ActDT to reflect procedure-specific patterns.

The contributions of this paper are three-fold.
First, we introduce text-to-video reasoning retrieval for OR clips, a task requiring fine-grained discrimination between visually near-identical clips through implicit queries that demand reasoning.
Correspondingly, we propose OR\textsuperscript{3}, which introduces ActDTs to encode temporal object interaction dynamics for OR clip retrieval.
Second, we introduce imagination-based retrieval to eliminate cross-modal semantic gaps, and evidence-grounded refinement to adapt query interpretation to procedure-specific patterns.
Third, we construct a benchmark for OR text-to-video reasoning retrieval.

\section{Methods}
\label{sec:method}

\begin{figure*}[t]
\centering
\includegraphics[width=0.85\linewidth]{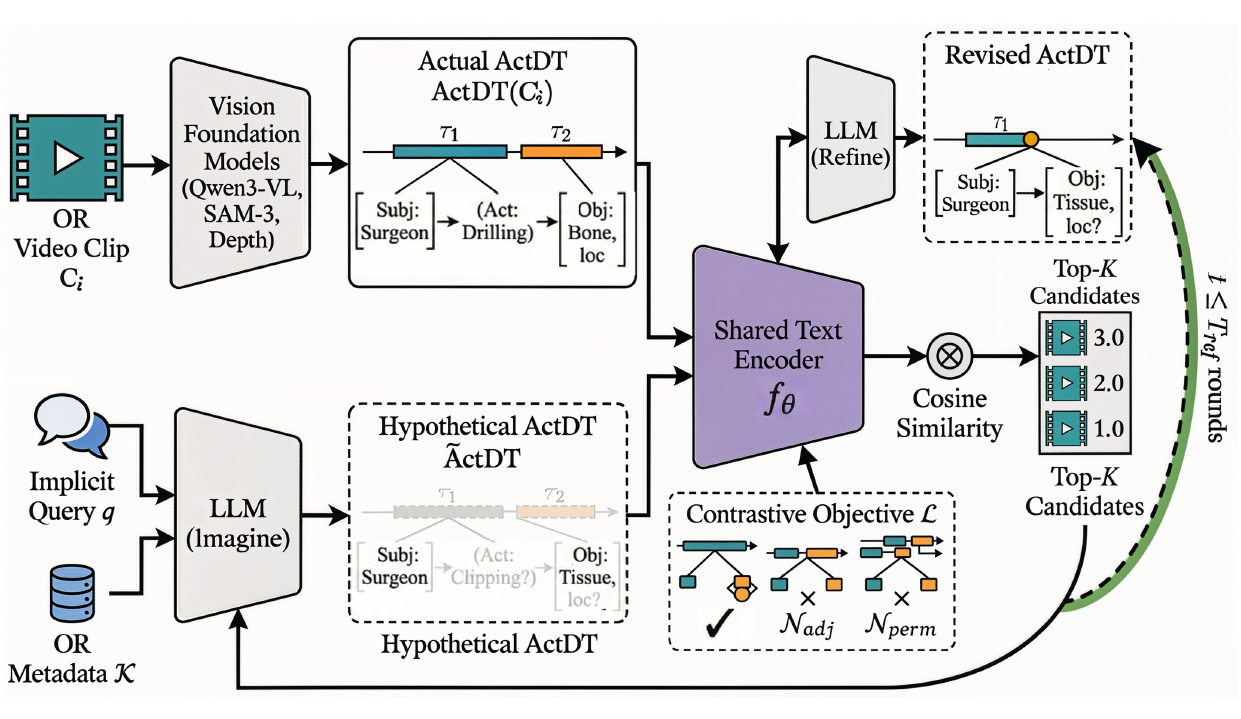}
\caption{Overview of OR\textsuperscript{3}.
Each OR video clip $C_i$ is converted into an actual ActDT through vision foundation models (Qwen3-VL, SAM-3, DepthAnythingV3), encoding subject-action-object triplets within non-overlapping temporal intervals.
Given an implicit query $q$ and OR metadata $\mathcal{K}$, an LLM imagines a hypothetical ActDT $\widehat{\text{ActDT}}_q$ that predicts the action primitives the target clip would contain.
A shared text encoder $f_\theta$, trained with a contrastive objective over adjacent-clip ($\mathcal{N}_{\text{adj}}$) and temporal-permutation ($\mathcal{N}_{\text{perm}}$) hard negatives~\cite{robinson2020contrastive}, encodes both representations and ranks clips by cosine similarity.
The top-$K$ candidates are then passed to an LLM that performs evidence-grounded refinement, revising the hypothetical ActDT across up to $T_{\text{ref}}$ rounds to align with procedure-specific patterns observed in the retrieved candidates.}
\label{fig:pipeline}
\end{figure*}

\subsubsection{Overview.}
\label{sec:formulation}
Consider an OR video segmented into an ordered clip sequence $\mathcal{C} = \{C_1, C_2, \ldots, C_M\}$ based on surgical phases and steps~\cite{mmor,czempiel2020tecno}.
Given an implicit text query $q$ that requires reasoning, our goal is to retrieve a ranked list of clips $\{(C_i, r_i)\}_{i=1}^{K}$, where $C_i \in \mathcal{C}$, $r_i \in [0, 1]$ denotes the relevance score, and $K$ is the number of clips returned~\cite{reasont2v}.
To address this, OR\textsuperscript{3} (shown in Fig.~\ref{fig:pipeline}) first converts each clip into an action-driven digital twin (ActDT) that encodes concurrent object interactions as temporally structured action primitives~\cite{dessalene2023therbligs} serialized in JSON.
Imagination-based retrieval then employs an LLM to generate a hypothetical ActDT from the query itself, and matches this imagined representation against actual clip ActDTs using a single text encoder.
Evidence-grounded refinement further revises the imagined ActDT based on discrepancies with top-ranked candidates, adapting to procedure-specific patterns.
The refined ActDT is re-encoded and matched against clip ActDTs to produce the final ranked results.

\subsubsection{Action-Driven Digital Twin.}
\label{sec:action_dt}
We represent each clip $C_i$ as an action-driven digital twin (ActDT), a sequence of action primitives that encode the object and their interactions.
Each ActDT is serialized as structured JSON text so that LLMs can process it directly without visual token compression of the raw video clips~\cite{shen2025temporally,shen2025operating,jit}.
Formally, ActDT can be defined as $\text{ActDT}(C_i) = [\, \{\, \texttt{interval}\!: \tau_k,\; \texttt{actions}\!: \big[\,\{\texttt{subj}\!: \mathbf{s}_{k,n},\; \texttt{act}\!: v_{k,n},\; \texttt{obj}\!: \mathbf{o}_{k,n}\}\,\big]_{n=1}^{N_k} \} \,]_{k=1}^{T_i}$,
where $T_i \in \mathbb{N}^+$ denotes the number of temporal intervals in clip $C_i$, and $N_k \in \mathbb{N}^+$ is the number of concurrent interactions within the $k$-th interval.
Each interval $\tau_k = [t_k^{\text{s}}, t_k^{\text{e}}] \in \mathbb{R}^2$ is defined by its start $t_k^{\text{s}}$ and end $t_k^{\text{e}}$ timestamps, with intervals not overlapping and their union spanning the entire duration of $C_i$.
The subject $\mathbf{s}_{k,n} = \big\{\, \texttt{cat}\!: c,\; \texttt{attr}\!: d,\; \texttt{loc}\!: \ell \,\big\}$ and object $\mathbf{o}_{k,n}$ share the same format, each containing three fields.
The category $c \in \mathcal{E}$ is drawn from an OR-specific vocabulary $\mathcal{E}$ of surgical roles, instruments, and equipment.
The attribute $d$ is a natural-language summary of properties such as posture or instrument state, generated by a VLM~\cite{qwen3vl} from the RGB frames of $C_i$.
The location $\ell \in \mathbb{R}^5$ encodes the bounding-box coordinates derived from the SAM-3~\cite{sam3} segmentation masks and the depth prediction in the object centroids of DepthAnythingV3~\cite{da3}.
The action $v_{k,n} \in \mathcal{V}$ records the interaction linking $\mathbf{s}_{k,n}$ to $\mathbf{o}_{k,n}$, also produced by the VLM and drawn from an OR-specific action vocabulary $\mathcal{V}$.

\subsubsection{Imagination-Based Retrieval.}
\label{sec:imagination}
OR\textsuperscript{3} reformulates retrieval as intra-modal matching rather than cross-modal alignment.
Given a query $q$, an LLM generates a hypothetical ActDT $\widehat{\text{ActDT}}_q = \text{LLM}_{\text{imagine}}(q, \mathcal{K})$, where $\mathcal{K}$ is procedural information drawn from OR video metadata.
The LLM does not rephrase $q$ but instead predicts the action primitives the target clip would contain.
The hypothetical ActDT shares the same interval-action structure as actual ActDTs but omits $\ell$ and replaces VLM-generated attributes $d$ with procedure-informed properties predicted by the LLM, as no visual frames are available at query time.
Because both hypothetical and actual ActDTs are serialized as JSON text, we encode them with a single text encoder $f_\theta$ and compute 
% $r_i = \mathrm{sim}\!\left(f_\theta(\widehat{\text{ActDT}}_q),\; f_\theta(\text{ActDT}(C_i))\right)$,
%
\begin{equation}
r_i = \mathrm{sim}\!\left(f_\theta(\widehat{\text{ActDT}}_q),\; f_\theta(\text{ActDT}(C_i))\right),
\end{equation}
where $\mathrm{sim}(\cdot, \cdot)$ denotes cosine similarity and $r_i$ is the relevance score for clip $C_i$.
We train $f_\theta$ with a contrastive objective combining two types of hard negatives tailored to ActDT.
The first type is temporally adjacent clips, which share nearly identical entities but differ in actions, forcing the encoder to attend to action-level distinctions.
The second type is temporal-permutation negatives, where a random permutation $\sigma$ reorders the intervals of each ActDT to produce $\text{ActDT}^{\sigma}(C_i) = [\text{ActDT}(C_i)_{\sigma(k)}]_{k=1}^{T_i}$, preserving action primitives while disrupting their temporal arrangement.
To bridge the structural gap between hypothetical and actual ActDTs, we apply field-dropout augmentation, independently masking each field with probability $p_{\text{drop}}$ to produce $\widetilde{\text{ActDT}}(C_i) = \mathcal{M}(\text{ActDT}(C_i), p_{\text{drop}})$, where $\mathcal{M}$ replaces selected fields with a null token.
This makes $f_\theta$ tolerant to the absence of the field in hypothetical ActDTs.
The training loss~\cite{oord2018representation,chen2020simple} over the positive pair $(q, C^+)$ where $C^+$ is the ground-truth clip for $q$, adjacent negatives $\mathcal{N}_{\text{adj}}(C^+)$ drawn from temporally neighboring clips, and permutation negatives $\mathcal{N}_{\text{perm}}(C^+)$ constructed by interval reordering is
\begin{equation}
\mathcal{L} = - \sum_{(q, C^+)} \log \frac{ \exp(r^+ / \gamma) }{ \exp(r^+ / \gamma) + \sum_{C^- \in \mathcal{N}_{\text{adj}} \cup \mathcal{N}_{\text{perm}}} \exp(r^- / \gamma) },
\end{equation}
where $\gamma$ is a temperature parameter, and $r^+ = \mathrm{sim}(f_\theta(\widehat{\text{ActDT}}_q),\, f_\theta(\widetilde{\text{ActDT}}(C^+)))$, $r^- = \mathrm{sim}(f_\theta(\widehat{\text{ActDT}}_q),\, f_\theta(\widetilde{\text{ActDT}}(C^-)))$.
The top-$K$ candidates ranked by $r_i$ proceed to evidence-grounded refinement.

\subsubsection{Evidence-Grounded Refinement.}
\label{sec:refinement}
The initial hypothetical ActDT relies on the LLM's generic knowledge, which may not capture the conventions of a specific procedure~\cite{reasont2v,jit}.
OR\textsuperscript{3} revises the imagined ActDT itself based on evidence from retrieved candidates.
After the initial ranking produces the top candidates $K$ with actual ActDTs $\{\text{ActDT}(C_{i_k})\}_{k=1}^{K}$, the LLM receives the original query $q$, the current hypothetical $\widehat{\text{ActDT}}_q$, and the candidate ActDTs, then identifies discrepancies between the imagined and observed action primitives.
Afterwards, the LLM generates a revised hypothetical, namely $\widehat{\text{ActDT}}_q^{\,(t+1)} = \text{LLM}_{\text{refine}}\!\left(q,\; \widehat{\text{ActDT}}_q^{\,(t)},\; \{\text{ActDT}(C_{i_k})\}_{k=1}^{K}\right)$,
%
% \begin{equation}
% \widehat{\text{ActDT}}_q^{\,(t+1)} = \text{LLM}_{\text{refine}}\!\left(q,\; \widehat{\text{ActDT}}_q^{\,(t)},\; \{\text{ActDT}(C_{i_k})\}_{k=1}^{K}\right),
% \end{equation}
%
where $t$ indexes the refinement round and $\widehat{\text{ActDT}}_q^{\,(0)} = \widehat{\text{ActDT}}_q$ is the initial imagined ActDT.
The revised hypothetical ActDT is re-encoded by $f_\theta$, and relevance scores $r_i$ are recomputed over $\mathcal{C}$ to produce an updated ranking.
This process iterates for at most $T_{\text{ref}}$ rounds or terminates early when the top-ranked clip remains unchanged.

\subsubsection{Benchmark.}
\label{sec:benchmark}
We construct an OR reasoning retrieval benchmark from the MM-OR dataset~\cite{mmor} to evaluate clip-level retrieval under implicit queries.
Each of the 17 full-length recordings is segmented into non-overlapping clips according to the annotated surgical phases and procedural steps, yielding 386 clips with durations ranging from 20 seconds to 10 minutes.
Two annotators with surgical domain knowledge independently come up implicit queries in four reasoning categories: temporal (\textit{e}.\textit{g}., ``\textit{the step right before bone sawing}''), causal (\textit{e}.\textit{g}., ``\textit{the action that triggered a sterility breach}''), procedural (\textit{e}.\textit{g}., ``\textit{the first use of the drill after robot calibration}''), and role-based (\textit{e}.\textit{g}., ``\textit{what the nurse prepared while the surgeon was sawing}'').
Each query is mapped to one or more ground-truth clips and cross-verified by the other annotator, with disagreements resolved through discussion to ensure unambiguous correspondence.
The final benchmark contains 276 implicit queries over 386 clips, split into training (162 queries, 218 clips), validation (48 queries, 72 clips), and test (66 queries, 96 clips) following the original MM-OR data partitions.

\section{Experiments}

\subsubsection{Implementation Details.}
We construct ActDTs using Qwen3-VL-8B~\cite{qwen3vl} as the VLM for generating action descriptions and entity attributes, SAM-3~\cite{sam3} for instance segmentation masks, and DepthAnythingV3~\cite{da3} for depth estimation at object centroids.
For imagination-based retrieval and evidence-grounded refinement, we employ Gemini-3-Pro as the LLM backbone, with top-$K{=}10$ candidates forwarded to refinement and a maximum of $T_{\text{ref}}{=}3$ refinement rounds.
The text encoder $f_\theta$ is initialized from the BERT-base model~\cite{bert} and trained with AdamW and learning rate $2{\times}10^{-5}$, weight decay $0.01$ for 80 epochs on a 8 NVIDIA GeForce RTX 4090 GPUs, with temperature $\gamma{=}0.07$, field-dropout probability $p_{\text{drop}}{=}0.3$, and two temporally adjacent clips on each side as hard negatives.
We report Recall at rank $K$ (R@$K$) for $K \in \{1, 5, 10\}$, measuring the fraction of queries for which at least one ground-truth clip appears within the top-$K$ results.
All metrics are averaged over five runs with different random seeds, and we report mean $\pm$ standard deviation.

\begin{table*}[t!]
\centering
\caption{Retrieval performance on the OR reasoning retrieval benchmark.
Methods are grouped into embedding-based retrieval, LLM approaches, reasoning retrieval, and our OR\textsuperscript{3}.
Bold and underline mark the best and second-best results.}
\label{tab:main_results}
\resizebox{\textwidth}{!}{
\begin{tabular}{lcccccccccccc|ccc}
\toprule
\multirow{2}{*}{Method} & \multicolumn{3}{c}{Temporal} & \multicolumn{3}{c}{Causal} & \multicolumn{3}{c}{Procedural} & \multicolumn{3}{c|}{Role-based} & \multicolumn{3}{c}{Overall} \\
\cmidrule(lr){2-4} \cmidrule(lr){5-7} \cmidrule(lr){8-10} \cmidrule(lr){11-13} \cmidrule(l){14-16}
& R@1 & R@5 & R@10 & R@1 & R@5 & R@10 & R@1 & R@5 & R@10 & R@1 & R@5 & R@10 & R@1 & R@5 & R@10 \\
\midrule
CLIP4Clip~\cite{clip4clip}
& 10.5{\tiny$\pm$2.4} & 26.3{\tiny$\pm$3.4} & 36.8{\tiny$\pm$3.8}
& 5.3{\tiny$\pm$1.9} & 15.8{\tiny$\pm$3.1} & 23.7{\tiny$\pm$3.6}
& 8.8{\tiny$\pm$2.1} & 20.6{\tiny$\pm$3.0} & 29.4{\tiny$\pm$3.4}
& 11.8{\tiny$\pm$2.6} & 26.5{\tiny$\pm$3.6} & 38.2{\tiny$\pm$3.9}
& 9.1{\tiny$\pm$1.8} & 22.7{\tiny$\pm$2.7} & 31.8{\tiny$\pm$3.0} \\
X-CLIP~\cite{xclip}
& 14.2{\tiny$\pm$2.7} & 31.6{\tiny$\pm$3.6} & 42.1{\tiny$\pm$3.8}
& 7.9{\tiny$\pm$2.3} & 18.4{\tiny$\pm$3.3} & 28.9{\tiny$\pm$3.8}
& 11.8{\tiny$\pm$2.4} & 23.5{\tiny$\pm$3.2} & 32.4{\tiny$\pm$3.5}
& 14.7{\tiny$\pm$2.9} & 32.4{\tiny$\pm$3.8} & 44.1{\tiny$\pm$4.0}
& 12.1{\tiny$\pm$2.1} & 25.8{\tiny$\pm$2.8} & 36.4{\tiny$\pm$3.1} \\
InternVideo2~\cite{internvideo2}
& 17.6{\tiny$\pm$3.0} & 36.8{\tiny$\pm$3.8} & 47.4{\tiny$\pm$3.9}
& 10.5{\tiny$\pm$2.6} & 23.7{\tiny$\pm$3.6} & 34.2{\tiny$\pm$4.0}
& 14.7{\tiny$\pm$2.6} & 32.4{\tiny$\pm$3.5} & 41.2{\tiny$\pm$3.7}
& 17.6{\tiny$\pm$3.1} & 38.2{\tiny$\pm$3.9} & 52.9{\tiny$\pm$4.0}
& 15.2{\tiny$\pm$2.3} & 33.3{\tiny$\pm$3.0} & 43.9{\tiny$\pm$3.2} \\
TeachCLIP~\cite{teachclip}
& 12.3{\tiny$\pm$2.5} & 28.1{\tiny$\pm$3.5} & 40.4{\tiny$\pm$3.8}
& 5.3{\tiny$\pm$1.9} & 15.8{\tiny$\pm$3.1} & 26.3{\tiny$\pm$3.7}
& 11.8{\tiny$\pm$2.4} & 23.5{\tiny$\pm$3.2} & 32.4{\tiny$\pm$3.5}
& 12.9{\tiny$\pm$2.7} & 29.4{\tiny$\pm$3.7} & 41.2{\tiny$\pm$4.0}
& 10.6{\tiny$\pm$2.0} & 24.2{\tiny$\pm$2.7} & 34.8{\tiny$\pm$3.1} \\
Qwen3-VL-8B~\cite{qwen3vl}
& 21.1{\tiny$\pm$3.2} & 42.1{\tiny$\pm$3.8} & 52.6{\tiny$\pm$3.9}
& 15.8{\tiny$\pm$3.1} & 28.9{\tiny$\pm$3.8} & 39.5{\tiny$\pm$4.1}
& 17.6{\tiny$\pm$2.8} & 35.3{\tiny$\pm$3.6} & 47.1{\tiny$\pm$3.7}
& 23.5{\tiny$\pm$3.4} & 44.1{\tiny$\pm$4.0} & 55.9{\tiny$\pm$4.0}
& 19.7{\tiny$\pm$2.5} & 37.9{\tiny$\pm$3.1} & 48.5{\tiny$\pm$3.2} \\
Gemini3
& 15.8{\tiny$\pm$2.8} & 33.3{\tiny$\pm$3.7} & 45.6{\tiny$\pm$3.9}
& 10.5{\tiny$\pm$2.6} & 21.1{\tiny$\pm$3.5} & 31.6{\tiny$\pm$3.9}
& 11.8{\tiny$\pm$2.4} & 29.4{\tiny$\pm$3.4} & 38.2{\tiny$\pm$3.7}
& 17.6{\tiny$\pm$3.1} & 35.3{\tiny$\pm$3.9} & 47.1{\tiny$\pm$4.0}
& 13.6{\tiny$\pm$2.2} & 30.3{\tiny$\pm$2.9} & 40.9{\tiny$\pm$3.2} \\
ReasonT2V~\cite{reasont2v}
& \underline{26.3}{\tiny$\pm$3.4} & \underline{47.4}{\tiny$\pm$3.9} & \underline{57.9}{\tiny$\pm$3.8}
& \underline{15.8}{\tiny$\pm$3.1} & \underline{31.6}{\tiny$\pm$3.9} & \underline{42.1}{\tiny$\pm$4.2}
& \underline{23.5}{\tiny$\pm$3.2} & \underline{41.2}{\tiny$\pm$3.7} & \underline{50.0}{\tiny$\pm$3.7}
& \underline{29.4}{\tiny$\pm$3.7} & \underline{47.1}{\tiny$\pm$4.0} & \underline{61.8}{\tiny$\pm$3.9}
& \underline{24.2}{\tiny$\pm$2.8} & \underline{42.4}{\tiny$\pm$3.2} & \underline{53.0}{\tiny$\pm$3.2} \\
\midrule
\textbf{OR\textsuperscript{3} (Ours)}
& \textbf{63.2}{\tiny$\pm$3.8} & \textbf{82.5}{\tiny$\pm$3.0} & \textbf{89.5}{\tiny$\pm$2.4}
& \textbf{42.1}{\tiny$\pm$4.2} & \textbf{63.2}{\tiny$\pm$4.1} & \textbf{73.7}{\tiny$\pm$3.7}
& \textbf{58.8}{\tiny$\pm$3.7} & \textbf{76.5}{\tiny$\pm$3.2} & \textbf{85.3}{\tiny$\pm$2.6}
& \textbf{67.6}{\tiny$\pm$3.8} & \textbf{85.3}{\tiny$\pm$2.9} & \textbf{91.2}{\tiny$\pm$2.3}
& \textbf{57.6}{\tiny$\pm$3.2} & \textbf{77.3}{\tiny$\pm$2.7} & \textbf{84.8}{\tiny$\pm$2.3} \\
\bottomrule
\end{tabular}%
}
\end{table*}

\begin{figure*}[b!]
\centering
\includegraphics[width=\linewidth]{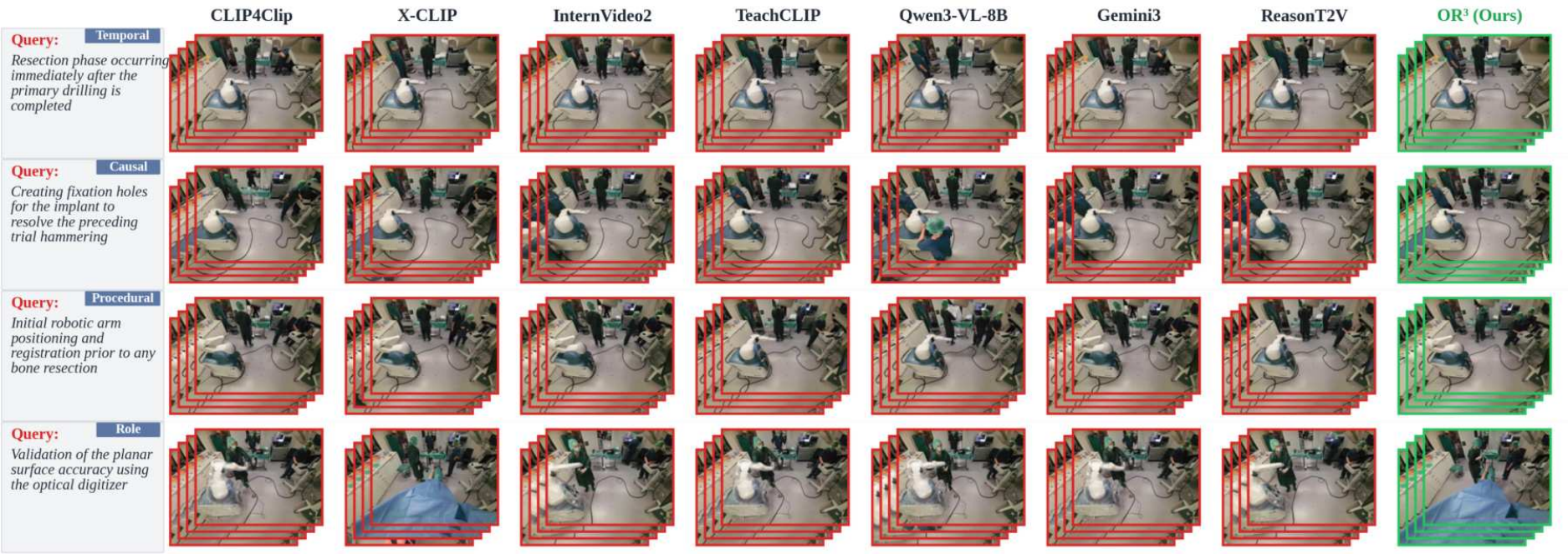}
\caption{Qualitative comparison on the OR reasoning retrieval benchmark across four query categories.
Red borders indicate incorrect retrievals where methods return clips that do not match the query, while green borders denote correct retrievals by OR\textsuperscript{3}.
%
% Embedding-based methods and video-LLMs retrieve visually similar but semantically wrong clips, as consecutive OR phases share near-identical appearances.
% %
% ReasonT2V improves over these baselines but still confuses clips whose object-centric digital twins overlap.
% %
% OR\textsuperscript{3} correctly identifies the target clip in each category by reasoning over action-level state transitions encoded in ActDTs.
}
\label{fig:qualitative}
\end{figure*}

\subsubsection{Retrieval Performance.}
\label{sec:exp_retrieval}
Table~\ref{tab:main_results} compares OR\textsuperscript{3} with embedding-based retrieval~\cite{clip4clip,xclip,teachclip}, LLM-based approaches~\cite{qwen3vl,internvideo2}, and ReasonT2V~\cite{reasont2v} on our OR reasoning retrieval benchmark.
OR\textsuperscript{3} achieves 57.6\% R@1, outperforming ReasonT2V~\cite{reasont2v} by 33.4 percentage points and confirming that ActDT captures action-level differences while imagination-based retrieval bridges the semantic gap between abstract implicit queries and digital twin representations.
Embedding-based methods~\cite{clip4clip,xclip,teachclip} remain below 16\% R@1, as global similarity matching cannot reason over implicit queries nor discriminate between visually near-identical OR clips differing only in actions.
LLM-based approaches improve through partial language reasoning, yet visual token compression~\cite{li2025fcot,reasont2v} discards fine-grained action and temporal structure needed for OR clip retrieval.
ReasonT2V~\cite{reasont2v} reaches 24.2\% R@1 through digital twin representations, but its object-centric design fails to capture action-level state transitions distinguishing consecutive surgical clips.
Across query categories, causal queries show most challenging, with OR\textsuperscript{3} achieving 42.1\% versus 15.8\% for ReasonT2V, as resolving cause-effect chains requires both temporal ordering and action-level reasoning.
Temporal and procedural queries show the largest absolute gains of 36.9 and 35.3 R@1 points over ReasonT2V, reflecting ActDT's interval-structured action primitives for queries referencing sequential ordering or workflow position.
Role-based queries yield the highest performance at 67.6\% R@1, since identifying agent-action pairs aligns naturally with the subject-action-object triplets in each ActDT interval.

\begin{table*}[t]
\centering
\caption{Ablation study on the OR reasoning retrieval benchmark (R@1).
ActDT: action-driven digital twin (replaced by object-centric digital twin of ReasonT2V when absent).
Imag.: imagination-based retrieval (replaced by cross-modal paired-encoder matching when absent).
Ref.: evidence-grounded refinement.
Adj.: adjacent-clip hard negatives.
Perm.: temporal-permutation negatives.
FD: field-dropout augmentation.}
\label{tab:ablation}
\resizebox{\textwidth}{!}{%
\begin{tabular}{cccccc|cccc|c}
\toprule
ActDT & Imag. & Ref. & Adj. & Perm. & FD & Temp. & Caus. & Proc. & Role & Overall \\
\midrule
\ding{55} & \checkmark & \checkmark & \checkmark & \checkmark & \checkmark & 47.4{\tiny$\pm$3.9} & 28.9{\tiny$\pm$3.8} & 41.2{\tiny$\pm$3.7} & 52.9{\tiny$\pm$4.0} & 42.4{\tiny$\pm$3.2} \\
\checkmark & \ding{55} & \checkmark & \checkmark & \checkmark & \checkmark & 50.9{\tiny$\pm$3.9} & 31.6{\tiny$\pm$3.9} & 44.1{\tiny$\pm$3.7} & 55.9{\tiny$\pm$4.0} & 45.5{\tiny$\pm$3.2} \\
\checkmark & \checkmark & \ding{55} & \checkmark & \checkmark & \checkmark & 54.4{\tiny$\pm$3.9} & 34.2{\tiny$\pm$4.0} & 50.0{\tiny$\pm$3.7} & 61.8{\tiny$\pm$3.9} & 50.0{\tiny$\pm$3.2} \\
\checkmark & \checkmark & \checkmark & \ding{55} & \checkmark & \checkmark & 57.9{\tiny$\pm$3.9} & 36.8{\tiny$\pm$4.1} & 50.0{\tiny$\pm$3.7} & 61.8{\tiny$\pm$3.9} & 51.5{\tiny$\pm$3.1} \\
\checkmark & \checkmark & \checkmark & \checkmark & \ding{55} & \checkmark & 59.6{\tiny$\pm$3.8} & 39.5{\tiny$\pm$4.1} & 55.9{\tiny$\pm$3.7} & 58.8{\tiny$\pm$3.9} & 53.0{\tiny$\pm$3.1} \\
\checkmark & \checkmark & \checkmark & \checkmark & \checkmark & \ding{55} & 56.1{\tiny$\pm$3.9} & 36.8{\tiny$\pm$4.1} & 55.9{\tiny$\pm$3.7} & 61.8{\tiny$\pm$3.9} & 52.3{\tiny$\pm$3.2} \\
\midrule
\checkmark & \checkmark & \checkmark & \checkmark & \checkmark & \checkmark & \textbf{63.2}{\tiny$\pm$3.8} & \textbf{42.1}{\tiny$\pm$4.2} & \textbf{58.8}{\tiny$\pm$3.7} & \textbf{67.6}{\tiny$\pm$3.8} & \textbf{57.6}{\tiny$\pm$3.2} \\
\bottomrule
\end{tabular}%
}
\end{table*}

\begin{figure*}[b!]
\centering
\includegraphics[width=\linewidth]{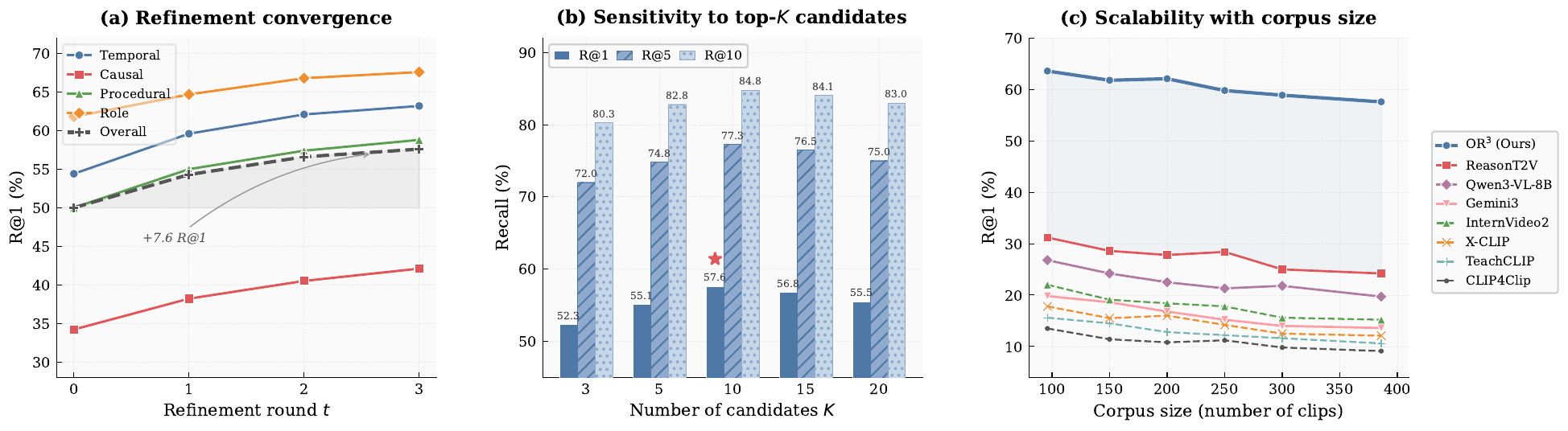}
\caption{Additional analyses of OR\textsuperscript{3}.
(a) R@1 per query category as a function of evidence-grounded refinement round $t$, where $t{=}0$ corresponds to the initial imagined ActDT before any refinement.
(b) Sensitivity to the number of top-$K$ candidates forwarded from imagination-based retrieval to refinement, with the star marking the selected operating point ($K{=}10$).
(c) R@1 of all compared methods as the retrieval corpus grows from 96 to 386 clips; the shaded region indicates the margin of OR\textsuperscript{3} over the strongest baseline.}
\label{fig:analysis}
\end{figure*}

Fig.~\ref{fig:qualitative} shows qualitative results in all four query categories.
Embedding-based methods~\cite{clip4clip,xclip,teachclip} and video-LLMs consistently retrieve visually similar but incorrect clips, as consecutive OR phases share nearly identical room layouts and personnel configurations.
ReasonT2V~\cite{reasont2v} narrows the gap through digital twin reasoning yet still confuses clips whose objects overlap but whose actions differ.
OR\textsuperscript{3} retrieves the correct clip in each case, confirming that action-driven digital twins and imagination-based matching together resolve the fine-grained distinctions that other approaches miss.

\subsubsection{Ablation Study.}
\label{sec:exp_ablation}
Table~\ref{tab:ablation} isolates each component by removing one at a time and reporting R@1 across query categories.
Replacing ActDT with ReasonT2V's object-centric digital twin representation~\cite{reasont2v} causes the largest drop of 15.2\% overall, with procedural queries falling 17.6\% and temporal queries 15.8\%, confirming that interval-structured action primitives drive fine-grained clip discrimination.
This aligns with the large performance gap in Table~\ref{tab:main_results} between OR\textsuperscript{3} and ReasonT2V~\cite{reasont2v} on temporal and procedural queries, where sequential action ordering determines relevance.
Eliminating imagination-based retrieval reduces overall R@1 by 12.1\%, validating that matching of the same-space eliminates the semantic gap between abstract queries and concrete ActDT representations.
Without evidence-grounded refinement, overall R@1 drops 7.6\%, with causal queries declining 7.9\%, as procedure-specific conventions matter most for resolving cause-effect chains.
Among training strategies, adjacent-clip hard negatives contribute 6.1\% by forcing the encoder to distinguish action-level differences between temporally neighboring clips.
Field-dropout augmentation adds 5.3\% by tolerating missing visual attributes in hypothetical ActDTs, while temporal-permutation negatives yield 4.6\% by preserving interval ordering.
All components produce additive gains exceeding their individual contributions, indicating mutual reinforcement.
Fig.~\ref{fig:analysis} presents additional analyses of refinement convergence, candidate pool sensitivity, and corpus scalability.
Evidence-grounded refinement yields monotonic R@1 gains across all query categories, with causal queries benefiting the most (34.2\% $\to$ 42.1\% over three rounds) and diminishing returns after $t{=}2$ supporting the early-stopping criterion.
Performance peaks at $K{=}10$ candidates, as smaller pools risk excluding the ground-truth clip while larger pools dilute the evidence set presented to the LLM.
As the corpus grows from 96 to 386 clips, OR\textsuperscript{3} loses only 6.0 R@1 points compared to 7.0 for ReasonT2V and up to half the initial accuracy for embedding-based methods, confirming that ActDT-based discrimination scales more gracefully with increasing visual ambiguity.

\section{Conclusion}
\label{sec:conclusion}
We present OR\textsuperscript{3}, a reasoning retrieval method that addresses the gap between how users search for OR clips and how the existing text-to-video retrieval method operates.
By shifting from object-centric to action-driven digital twins, our approach captures the temporal dynamics and state transitions that distinguish visually similar OR clips.
Imagination-based retrieval further reframes cross-modal matching as an intra-modal problem, while evidence-grounded refinement adapts query interpretation to the conventions of individual procedures.
Future work can explore evaluation of other procedure types and larger clip collections.
Extending the framework to support multi-turn queries and real-time retrieval during live procedures can be another promising future direction.

\subsubsection{Acknowledgments.}
Y. Shen was supported in part by the JHU Amazon Initiative for Artificial Intelligence (AI2AI) fellowship program.

\subsubsection{Disclosure of Interests.}
The authors have no competing interests in the paper.

\bibliographystyle{plain}
\bibliography{ref}

\end{document}